\documentclass[10pt,twocolumn,letterpaper]{article}

\usepackage{cvpr}
\usepackage{times}
\usepackage{epsfig}
\usepackage{graphicx}
\usepackage{amsmath}
\usepackage{amssymb}
\usepackage{adjustbox}
\usepackage{array}
\usepackage{multirow}
\usepackage[sort,nocompress]{cite}
\newcommand*{\pd}[3][]{\ensuremath{\frac{\partial^{#1} #2}{\partial #3}}}

\usepackage[pagebackref=true,breaklinks=true,letterpaper=true,colorlinks,bookmarks=false]{hyperref}

 \cvprfinalcopy 
\def\cvprPaperID{6020} 

\ifcvprfinal\fi
\begin{document}
\title{Object Counting and Instance Segmentation with Image-level Supervision}
\author{Hisham  Cholakkal \thanks{Equal contribution} \and Guolei Sun \footnotemark[1] \and Fahad Shahbaz Khan \and Ling Shao \and \\
Inception Institute of Artificial Intelligence, UAE\\
{\tt\small {firstname.lastname}@inceptioniai.org}
}


\maketitle

\begin{abstract}
  Common object counting in a natural scene is a challenging problem in computer vision with numerous real-world applications. Existing image-level supervised common object counting approaches only predict the global object count and rely on additional instance-level supervision to also determine object locations. We propose an image-level supervised approach that provides both the global object count and the spatial distribution of object instances by constructing an object category density map. Motivated by psychological studies, we further reduce image-level supervision using a limited object count information (up to four). To the best of our knowledge, we are the first to propose image-level supervised density map estimation for common object counting and demonstrate its effectiveness in image-level supervised instance segmentation.  Comprehensive experiments are performed on the PASCAL VOC and COCO datasets. Our approach outperforms existing methods, including those using instance-level supervision, on both datasets for common object counting. Moreover, our approach  improves state-of-the-art image-level supervised instance segmentation \cite{PRM} with a relative gain of 17.8$\%$ in terms of average best overlap, on the PASCAL VOC 2012 dataset. \footnote{Code is publicly available at \href {https://github.com/GuoleiSun/CountSeg}{github.com/GuoleiSun/CountSeg}}

\end{abstract}

\section{Introduction}
\begin{figure}[t]
		
			\includegraphics[width=1\linewidth, clip=true, trim=0cm 14.1cm 11.6cm 0cm]{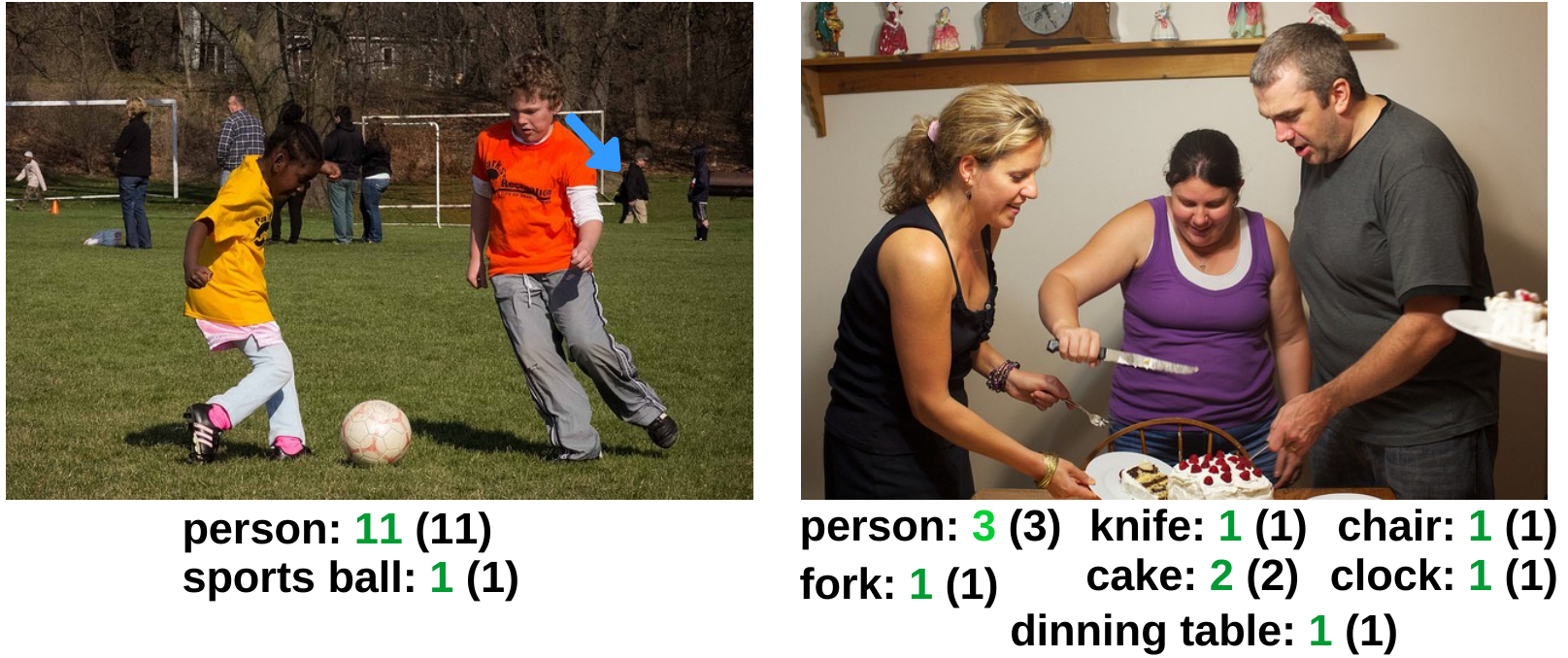}\\  \vspace{-0.30 cm}
			
			\caption{Object counting on COCO dataset. The ground-truth and our predictions are shown in black and green, respectively. Despite being trained using image-level object counts within the subitizing range [1-4], it accurately counts objects beyond the subitizing range (11 persons) under heavy occlusion (marked with blue arrow to show two persons) in the left image and diverse object categories in the right.
			}
			\label{fig:CountingIntro}
\end{figure}

\begin{figure*}[t]
		\centering
			\includegraphics[width=0.9\textwidth, clip=true, trim=0cm 18cm 7.2cm 0cm]{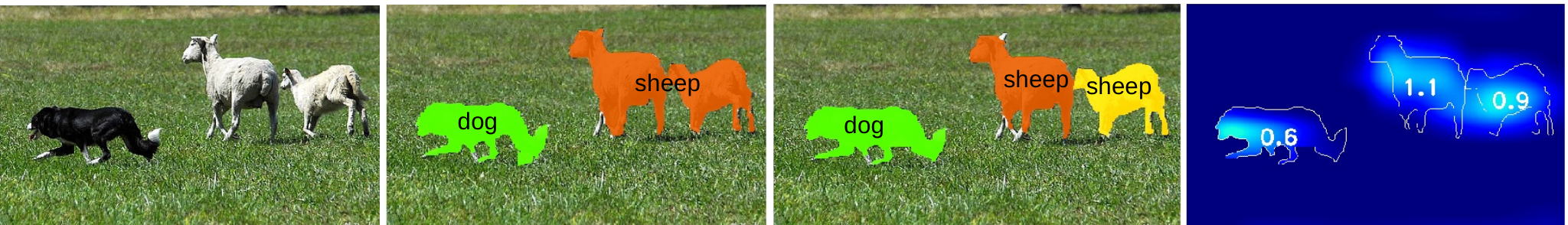}\\
				\centering
				\includegraphics[width=0.9\textwidth, clip=true, trim=0cm 14cm 7.2cm 3.2cm]{images/introduction_segmentation3.pdf}\\
				\vspace{-0.5cm}
			\hspace*{0.01\linewidth} (a) Input Image \hspace*{0.1\linewidth} (b) PRM \cite{PRM}\hspace*{0.1\linewidth}(c) Our Approach  \hspace*{0.07\linewidth} (d) Our Density Map \\  
			
			\caption{Instance segmentation examples using the PRM method \cite{PRM} (b) and our approach (c), on the PASCAL VOC 2012. Top row: The PRM approach \cite{PRM} fails to delineate spatially adjacent two sheep category instances. Bottom row: single person parts predicted as multiple persons along with inaccurate mask separation results in over-prediction (7 instead of 5). Our approach produces accurate masks by exploiting the spatial distribution of object count in per-category density maps (d). Density map accumulation for each predicted mask is shown inside the contour drawn for clarity. In the top row, density maps for sheep and dog categories are overlaid. 
			}
			\label{fig:SegmentIntro}
			\vspace*{-0.3cm}
\end{figure*}

 Common object counting, also referred as \textit{generic object counting}, is the task of accurately predicting the number of different object category instances present in natural scenes (see Fig.~\ref{fig:CountingIntro}).  The common object categories in natural scenes can vary from fruits to animals and the counting must be performed  in both indoor and outdoor  scenes (\eg COCO or PASCAL VOC datasets). 
 Existing works employ a localization-based strategy or utilize regression-based models directly optimized to predict object count, where the latter has been shown to provide superior results \cite{Chattopadhyay_2017_CVPR}. However, regression-based methods only predict the global object count without determining object locations. Beside global counts, the spatial distribution of objects in the form of a  per-category density map is helpful in other tasks, e.g., to delineate adjacent objects in instance segmentation  (see Fig. \ref{fig:SegmentIntro}).

 The problem of density map estimation to preserve the spatial distribution of people is well studied in crowd counting \cite{Cao_2018_ECCV,FullyConvDensityICCV17, Zhang_2016_CVPR, CSRnetDialatedConv_2018_CVPR,rankingUnlabelleddata_CVPR2018}. Here, the global count for the image is obtained
by summing over the predicted density map. Standard crowd density map estimation methods are required to predict large number of person counts in the presence of occlusions, e.g., in surveillance applications. The key challenges of constructing a density map in natural scenes are different to those in crowd density estimation, and include large intra-class variations in generic objects, co-existence of multiple instances of different objects in a scene (see Fig.~\ref{fig:CountingIntro}), and sparsity due to many objects having zero count on multiple images. 


Most methods for crowd density estimation use instance-level (point-level or bounding box) supervision  that requires manual annotation of each instance location.  Image-level supervised training alleviates the need for such user-intensive annotation by requiring only the count of different object instances in an image. 
We propose an image-level supervised density map estimation approach for natural scenes, that predicts the global object count while preserving the spatial distribution of objects.

Even though image-level supervised object counting reduces the burden of human annotation and is much weaker compared to instance-level supervisions, it still requires each object instance to be counted sequentially.
 Psychological studies \cite{psychological_study1,psychological_study2,psychological_study3,psychological_study4} have suggested that humans are capable of counting objects non-sequentially using holistic cues for fewer object counts, termed as a subitizing range (generally 1-4). We utilize this property to further reduce image-level supervision by only using object count annotations within the subitizing range. For short, we call this image-level lower-count (ILC) supervision. Chattopadhyay \etal\cite{Chattopadhyay_2017_CVPR} also investigate common object counting, where  object counts (both within and beyond the subitizing range) are used to predict the global object count. Alternatively, instance-level (bounding box) supervision is used to count objects by dividing an image into non-overlapping regions, assuming each region count falls within the subitizing range. Different to these strategies \cite{Chattopadhyay_2017_CVPR}, our ILC supervised approach requires neither bounding box annotation nor information beyond the subitizing range to predict both the  count and the spatial distribution of object instances.

In addition to common object counting, the proposed ILC supervised density map estimation is suitable for other scene understanding tasks. Here, we investigate its effectiveness for image-level supervised instance segmentation, where the task is to localize each object instance with pixel-level accuracy, provided image-level category labels. Recent work of \cite{PRM}, referred as peak response map (PRM), tackles the problem by boosting the local maxima (peaks) in the class response maps \cite{oquab2015object} of an image classifier using a peak stimulation module. A scoring metric is then used to rank off-the-shelf object proposals \cite{mcg_2017,cob_eccv2016} corresponding to each peak for instance mask prediction.  However, PRM  struggles to delineate spatially adjacent object instances from the same object category (see Fig. \ref{fig:SegmentIntro}(b)). We introduce a penalty term into the scoring metric that assigns a higher score to object proposals with a predicted count of~$1$, providing improved results (Fig. \ref{fig:SegmentIntro}(c)). The predicted count is obtained by accumulating the density map over the entire object proposal region (Fig. \ref{fig:SegmentIntro}(d)). 

 \noindent\textbf{Contributions:} We propose an ILC supervised density map estimation approach for common object counting. A novel loss function is introduced to construct per-category density maps with explicit terms for predicting the global count and spatial distribution of objects.  
 We also demonstrate the applicability of the proposed approach for image-level  supervised instance segmentation. For common object counting, our ILC supervised approach outperforms state-of-the-art instance-level supervised methods with a relative gain of 6.4$\%$ and 2.9$\%$, respectively, in terms of mean root mean square error (mRMSE), on the PASCAL VOC 2007 and COCO datasets. For image-level supervised instance segmentation, our approach  improves the state of the art from 37.6 to 44.3 in terms of average best overlap (ABO), on the PASCAL VOC 2012 dataset.
  	\begin{figure*}[t]
		\centering
						\includegraphics[width=0.97\linewidth, clip=true, trim=0cm 15.5cm 13cm 0cm]{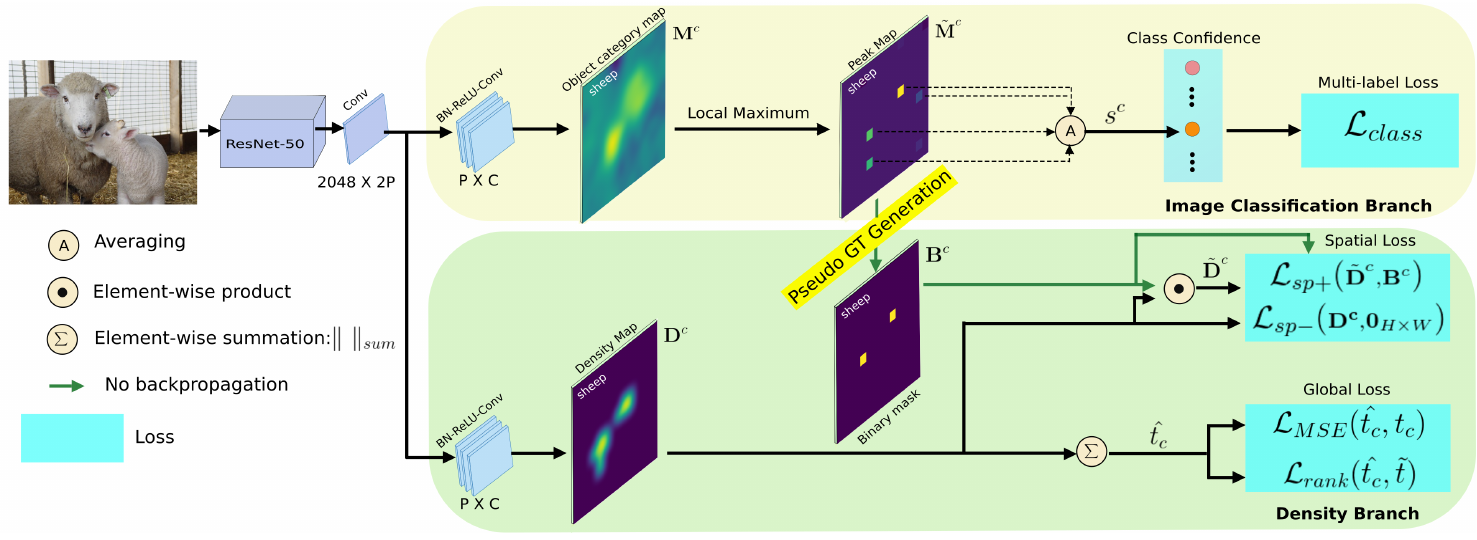}\\ \vspace{-0.1cm}
			
			\caption{Overview of our overall architecture. Our network has an image classification and a density branch, trained jointly using ILC supervision. The image classification branch predicts the presence and absence of objects. This branch is used to generate pseudo ground-truth for training the density branch. The density branch has two terms (spatial and global) in the loss function and produces a density map to predict the global object count and preserve the spatial distribution of objects. }
			\label{Fig:architectue}
			\vspace*{-0.2cm}
\end{figure*}

\vspace{-0.15cm}
\section{Related work}



Chattopadhyay \etal\cite{Chattopadhyay_2017_CVPR} investigated regression-based common object counting, using image-level (per-category count) and instance-level (bounding box) supervisions. The image-level supervised strategy, denoted as glancing, used count annotations from both within and beyond the subitizing range to predict the global count of objects, without providing information about their location. The instance-level (bounding box) supervised strategy, denoted as subitizing, estimated a large number of objects by dividing an image into non-overlapping regions, assuming the object count in each region falls within the subitizing range. Instead, our ILC supervised approach requires neither bounding box annotation nor beyond subitizing range count information during training. It then predicts the global object count, even beyond the subitizing range, together with the spatial distribution of object instances. 

 Recently, Laradji \etal \cite{WhereAreBlobsECCV18} proposed a localization-based counting approach, trained using instance-level (point) supervision\cite{point_feifei}. During inference, the model outputs blobs indicating the predicted locations of objects of interest and uses \cite{connectedComponent} to estimate object counts from these blobs. Different to \cite{WhereAreBlobsECCV18}, our approach is image-level supervised and directly predicts the object count through a simple summation of the density map without requiring any post-processing \cite{connectedComponent}.
 Regression-based  methods  generally perform well in the presence of occlusions \cite{Chattopadhyay_2017_CVPR, OxfordDensityNIPS2010}, while 
 localization-based counting approaches \cite{WhereAreBlobsECCV18,Girshick15ICCV} generalize well with a limited number of training images \cite{OxfordDensityNIPS2010,WhereAreBlobsECCV18}. Our method aims to combine the advantages of both approaches through a novel loss function that jointly optimizes the network to predict object locations and global object counts in a density map. 

   Reducing object count supervision for salient object subitizing was investigated in \cite{sos_subitizing_cvpr2015}. 
 However, their task is class-agnostic and subitizing is used to only count within the subitizing range. Instead, our approach constructs category-specific density maps and accurately predicts object counts both within and beyond the subitizing range. Common object counting has been previously used to improve object detection \cite{Chattopadhyay_2017_CVPR,Gao_2018_ECCV}. Their approach only uses the count information during detector training with no explicit component for count prediction. In contrast, our approach explicitly learns to predict the global object count. 
\vspace{-0.2cm}
\section{Proposed method}
\vspace{-0.1cm}
\label{proposedmethod_start}
Here, we present our image-level lower-count (ILC) supervised density map estimation approach. 
Our approach is built upon an ImageNet pre-trained network backbone (ResNet50) \cite{ResNet}. The proposed network architecture has two output branches: image classification and density branch (see Fig.~\ref{Fig:architectue}). The image classification branch estimates the presence or absence of objects, whereas the density branch predicts the global object count and the spatial distribution of object instances by constructing a density map. We remove the global pooling layer from the backbone and adapt the fully connected layer with a $1\times1$ convolution having $2P$ channels as output. We  divide these $2P$ channels equally between the image classification and density branches. We then add a $1\times 1$ convolution having $C$ output channels in each branch, resulting in a fully convolutional network \cite{FCN}. Here, $C$ is the number of object categories and $P$ is empirically set to be proportional to $C$. In each branch, the convolution is preceded by a batch normalization and a ReLU layer. The first branch provides object category maps and the second branch produces a density map for each object category.
\vspace{-0.1cm}
\subsection{The Proposed Loss Function}
\vspace{-0.1cm}
Let $\textbf{I}$ be a training image and $\textbf{t}=\{t_1, t_2, ..., t_c, ..., t_C\}$ be the corresponding vector for the ground-truth count of $C$~object categories. Instead of using an absolute object count, we employ a lower-count strategy to reduce the amount of image-level supervision. Given an image $\textbf{I}$, object categories are divided into three  non-overlapping sets based on their respective instance counts. The first set, $A$, indicates object categories which are absent in $\textbf{I}$ (i.e., $t_c=0$). The second set, $S$, represents categories within the subitizing range (i.e, $0<t_c \le 4$). The final set, $\tilde{S}$, indicates categories beyond the subitizing range (i.e, $t_c\ge \tilde{t}$, where $\tilde{t}=5$).





Let $\textbf{M}=\{\mathbf{M}^1, \mathbf{M}^2, ..., \mathbf{M}^c, ..., \mathbf{M}^C\}$
 denote the object category maps in the image classification branch, where $\mathbf{M}^c \in R ^{ H\times W}$. Let $\textbf{D}=\{\mathbf{D}^1, \mathbf{D}^2, ..., \mathbf{D}^c, ..., \mathbf{D}^C\}$ represent density maps produced by the density branch, where $\mathbf{D}^c \in R ^{ H\times W}$. Here,
 $H\times W$ is the spatial size of both the object category and density maps. The image classification and density branches are jointly trained, in an end-to-end fashion, given only ILC supervision with the following loss function:
 \vspace{-0.1cm}
\begin{equation} 
\label{eq:maineq}
     \mathcal{L}=  {{\cal{L}}_{class}}+\underbrace{{\cal{L}}_{spatial}+{\cal{L}}_{global}}_{Density~map~branch}.
      \vspace{-0.1cm}
\end{equation}
Here, the first term refers to  multi-label image classification loss \cite{multilabelsoftmargin} (see Sec.~\ref{peakstimulation}).  The last two terms, ${\cal{L}}_{spatial}$ and ${\cal{L}}_{global}$, are used to train the density branch (Sec.~\ref{density}). 
\vspace{-0.3cm}
\subsubsection{Image Classification Branch}
\label{peakstimulation}
Generally, training a density map requires instance-level supervision, such as point-level annotations \cite{OxfordDensityNIPS2010}. Such information is unavailable in our ILC supervised setting. To address this issue, we propose to generate pseudo ground-truth by exploiting the coarse-level localization capabilities of an image classifier \cite{oquab2015object, CAM} via object category maps.  These object category maps are generated from a fully convolutional architecture shown in Fig.~\ref{Fig:architectue}. 

While specifying classification confidence at each image location, class activation maps (CAMs) struggle to delineate multiple instances from the same object category \cite{oquab2015object, CAM}. Recently, the local maxima of CAMs are further boosted, to produce object category maps, during an image-classifier training for image-level supervised instance segmentation \cite{PRM}. Boosted local maxima  aim at falling on distinct object instances. For details on boosting local maxima, we refer to \cite{PRM}. Here, we use local maxima  locations to generate pseudo ground-truth for training the density branch.


As described earlier, object categories in $\textbf{I}$ are divided into three non-overlapping sets: $A$, $S$ and $\tilde{S}$. To train a one-versus-rest image classifier, we derive
binary labels from $t_c$ that indicate the presence $\forall c\in\{S, \tilde{S}\}$ or absence  $\forall c \in A$ of object categories. Let $\tilde {\textbf{M}}^{c}  \in R ^{H\times W}$ be the peak map derived from $c^{th}$ object category map ($\textbf{M}^{c}$) of $\textbf{M}$ such that:
 \vspace{-0.1cm}
\begin{equation*}
\small
 \tilde{\textbf{M}}^{c}(i,j)=\begin{cases}
    \textbf{M}^{c}(i,j),&\text{if $\textbf{M}^{c}(i,j)>\textbf{M}^{c}(i-r_i,j-r_j)$},\\
    0, & \text{otherwise}.
  \end{cases}
   \vspace{-0.0cm}
\end{equation*}
Here,  $-r \leq r_i \leq r$, ~  $-r \leq r_j \leq  r$ where  $r$ is the radius for the local maxima computation. We set $r=1$, as in \cite{PRM}. The local maxima are searched at all spatial locations with a stride of one. 
To train an image classifier, a class confidence score $s^c$ of the $c^{th}$ object category is computed as the average of non-zero elements of $\tilde{\textbf{M}}^{c}$. In this work, we use the multi-label soft-margin loss \cite{multilabelsoftmargin} for binary classification. 
 \vspace{-0.35cm}
\subsubsection{Density Branch}
\label{density} \vspace{-0.15cm}
 The image classification branch described above predicts the presence or absence of objects by using the class confidence scores derived from the peak map $\tilde {\textbf{M}}^{c}$. However, it struggles to differentiate between multiple objects and single object parts due to the lack of prior information about the number of object instances (see Fig.~\ref{fig:SegmentIntro}(b)). This causes a large number of false positives in the peak map $\tilde{\textbf{M}}^{c}$. Here, we utilize the count information and introduce a pseudo ground-truth generation scheme that prevents training a density map at those false positive locations. 
 
 When constructing a density map, it is desired to estimate accurate object counts at any image sub-region. Our spatial loss term ${\cal{L}}_{spatial}$ in Eq.~\ref{eq:maineq} ensures that individual object instances are localized while the global term ${\cal{L}}_{global}$ constrains the global object count to that of the ground-truth. This enables preservation of the spatial distribution of object counts in a density map. Later, we show that this property helps to improve instance segmentation.
 
 
 
 
 

  
  \noindent\textbf{Spatial Loss:} 
The spatial loss  ${\cal{L}}_{spatial}$ is divided into the loss $\mathcal{L}_{sp+}$ which enhances the positive peaks corresponding to instances of object categories within ${S}$, and the loss $\mathcal{L}_{sp-}$ which suppresses false positives of categories within ${A}$.   
 Due to the unavailability of absolute object count, the set $\tilde{S}$ is not used in the spatial loss and treated separately later. To enable ILC supervised density map training using  ${\cal{L}}_{spatial}$, we generate a pseudo ground-truth binary mask from peak map $\tilde{\textbf{M}}^{c}$.

  

  
   
   
 \noindent{\textit{Pseudo Ground-truth Generation:}} To compute the spatial loss $\mathcal{L}_{sp+}$, a pseudo ground-truth is generated for set $S$. For all object categories $c \in S$, the ${t_c}$-{th} highest peak value of peak map $\tilde{M}^{c}$ is computed using the heap-max algorithm \cite{max_heap}. The ${t_c}$-{th} highest peak value $h_{c}$ is then used to generate a pseudo ground-truth binary mask $\textbf{B}^{c}$ as,
 \vspace{-0.1cm}
\begin{equation} 
    \textbf{B}^{c}=u(\tilde{\textbf{M}}^{c}-h_c).
     \vspace{-0.1cm}
\end{equation}
 Here, $u(n)$ is the unit step function which is $1$ only if $n \ge 0$. Although the non-zero elements of the pseudo ground-truth mask $\textbf{B}^{c}$ indicate object locations, its zero elements do not necessarily point towards the background. Therefore, we construct a masked density map $\tilde{\textbf{D}}^{c}$ to exclude density map $\textbf{D}^{c}$ values at locations where the corresponding $\textbf{B}^{c}$ values are zero. Those density map $\textbf{D}^{c}$ values should also be excluded during the loss computation in Eq.~\ref{eq:possptial} and backpropagation (see Sec.~\ref{sec:backpropagate}), due to their risk of introducing false negatives. This is achieved by computing the Hadamard product between the density map $\textbf{D}^{c}$ and $\textbf{B}^{c}$ as,
 \vspace{-0.1cm}
\begin{equation} 
\label{eq:hadmard}
    \tilde{\textbf{D}}^{c}=\textbf{D}^{c} \odot \textbf{B}^{c}.
    \vspace{-0.1cm}
\end{equation}
The spatial loss $\mathcal{L}_{sp+}$ for object categories within the subitizing range ${S}$ is computed between $\textbf{B}^{c}$ and $\tilde{\textbf{D}}^{c}$ using a logistic binary cross entropy (logistic BCE) \cite{pytorch_cite} loss for positive ground-truth labels. 
 The logistic BCE loss transfers the network prediction ($\tilde{\textbf{D}}^{c}$) through a  sigmoid activation  layer $\sigma$ and computes the standard BCE loss as,
 \vspace{-0.1cm}
   \begin{equation} 
    \label{eq:possptial}
  \mathcal{L}_{sp+}(\tilde{\textbf{D}}^{c},~\textbf{B}^{c}) = - \sum_{\forall c \in S} \frac{\|\textbf{B}^{c} \odot\log ( \sigma(\tilde{\textbf{D}}^{c}))\|_{sum}} {|S|\cdot\|\mathbf{B}^{c}\|_{sum}}.
   \vspace{-0.1cm}
\end{equation} 
Here, $|S|$ is the cardinality of the set $S$ and the norm $\|~\|_{sum}$ is computed by taking the summation over all elements in a matrix. For example,  $\|\mathbf{B}^{c}~\|_{sum}$ = $\mathbf{1}^h \mathbf{B}^{c} \mathbf{1}^w$, where $\mathbf{1}^h$ and $\mathbf{1}^w $ are all-ones vectors of size ${1\times H}$  and ${W\times 1}$, respectively.
Here, the highest $t_c$ peaks in $\tilde{\textbf{M}}^{c}$ are assumed to fall on $t_c$  instances of object category $c \in S$. 
Due to the unavailability of ground-truth object locations, we use this assumption and observe that it holds in most scenarios.

The spatial loss $\mathcal{L}_{sp+}$ for the positive ground-truth labels enhances positive peaks corresponding to instances of object categories within ${S}$. However, the false positives of the density map for $c \in S$ are not penalized in this loss. We therefore introduce another term,  $\mathcal{L}_{sp-}$, into the loss function to address the false positives of $c\in A$.
For  $c\in A$,  positive  activations of $\textbf{D}^{c}$ indicate false detections. A zero-valued mask ${\mathbf{0_{H\times W}}}$ is used as ground-truth to reduce such false detections using logistic BCE loss,
 \vspace{-0.1cm}
    \begin{equation} 
  \mathcal{L}_{sp-}(\mathbf{D}^c,\mathbf{0}_{H\times W}) = -\sum_{c\in A}{{ \frac{\|\log (1- \sigma({\textbf{D}}^{c})\|_{sum} }{|A|\cdot H\cdot W}}}.
   \vspace{-0.1cm}
\end{equation}
Though the spatial loss ensures the preservation of spatial distribution of objects, only relying on local information may result in deviations in the global object count. 



\noindent\textbf{Global Loss:}
The global loss penalizes the deviation of the predicted count $\hat{t_c}$ from the ground-truth. It has two components: ranking loss $\mathcal{L}_{rank}$ for object categories beyond the subitizing range (i.e., $\forall c\in \tilde{S}$) and mean-squared error (MSE) loss $\mathcal{L}_{MSE}$ for the rest of the categories.  
$\mathcal{L}_{MSE}$ penalizes the predicted density map, if the global count prediction does not match with the ground-truth count. i.e., 
 \vspace{-0.1cm}
\begin{equation} 
\mathcal{L}_{MSE}(\hat{t_c},t_c)= \sum_{c\in \{A,S\}}\frac{({\hat{t_c}-t_c)^2 }}{\small{|A|+|S|}}.
 \vspace{-0.1cm}
 \end{equation}
 Here, the predicted count $\hat{t_c}$ is the  accumulation of the density map for a category $c$ over its entire spatial region. \ie 
  $\hat{t_c}=\|\textbf{D}^{c}\|_{sum}$. 
Note that object categories in $\tilde{S}$ were not previously considered in the computation of spatial loss $\mathcal{L}_{spatial}$ and mean-squared error loss $\mathcal{L}_{MSE}$. Here, we introduce a ranking loss \cite{rankingloss_cvpr2014} with a zero margin that penalizes under-counting for object categories within $\tilde{S}$,
 \vspace{-0.1cm}
\begin{equation} 
\mathcal{L}_{rank}(\hat{t_c}, \tilde{t})=\sum_{c \in \tilde{S}}{\!{\frac{max(0, \tilde{t}-{\hat{t_c}})}{|\tilde{S}|}}}.
 \vspace{-0.1cm}
 \end{equation}
The ranking loss  penalizes the density branch if the predicted object count $\hat{t_c}$ is less than $\tilde{t}$ for $c\in \tilde{S}$. Recall, the beyond subitizing range $\tilde{S}$ starts from $\tilde{t}=5$.

Within the subitizing range $S$, the spatial loss term $\mathcal{L}_{spatial}$ is optimized to locate object instances  while the global MSE loss ($\mathcal{L}_{MSE}$) is optimized for accurately predicting the corresponding global count. Due to the joint optimization of both these terms within the subitizing range, the network learns to correlate between the located objects and the global count. Further, the network is able to locate object instances, generalizing beyond the subitizing range $\tilde{S}$ (see Fig.~\ref{fig:SegmentIntro}). Additionally, the ranking loss $\mathcal{L}_{rank}$ term in the proposed loss function ensures the penalization of  under counting beyond the subitizing range $\tilde{S}$.

\noindent\textbf{Mini-batch Loss:}Normalized loss terms $\mathcal{\hat{L}}_{sp+}$, $\mathcal{\hat{L}}_{sp-}$, $\mathcal{\hat{L}}_{MSE}$ and $\mathcal{\hat{L}}_{rank}$ are computed by averaging respective loss terms over  all  images in the mini-batch. The  $\mathcal{L}_{spatial}$ is computed by $\mathcal{\hat{L}}_{sp+}+  \mathcal{\hat{L}}_{sp-}$.  For categories beyond the subitizing range,  $\mathcal{\hat{L}}_{rank}$ can lead to over-estimation of the count. Hence,  $\mathcal{L}_{global}$ is computed by  assigning  a relatively lower weight ($\lambda=0.1$) to $\mathcal{\hat{L}}_{rank}$ (see Table.~\ref{tab:loss_analysis}). \ie,  $ \mathcal{L}_{global}=\mathcal{\hat{L}}_{MSE}+  \lambda* \mathcal{\hat{L}}_{rank}$.
\vspace{-0.2cm}
\subsection{Training and Inference}
\vspace{-0.1cm}
\label{sec:backpropagate}
Our network is trained in two stages. In the first stage, the density branch is trained with only $\mathcal{L}_{MSE}$ and $\mathcal{L}_{rank}$ losses using  $S$ and $\tilde{S}$ respectively.  
The spatial loss $\mathcal{L}_{spatial}$ in Eq.~\ref{eq:maineq} is excluded in the first stage,  since it requires a pseudo ground-truth generated from the image classification branch.  The second stage includes the spatial loss. 

\noindent\textbf{Backpropagation:}
We use $\textbf{B}^{c}$ derived from the image classification branch as a pseudo ground-truth to train the density branch. Therefore, the backproapation of gradients through $\textbf{B}^{c}$ to the classifier branch is not required (shown with green arrows in Fig. \ref{Fig:architectue}).  The image classification branch is backpropagated as in \cite{PRM}. 
 In the density branch,  we use Hadamard product of the density map with  $\textbf{B}^{c}$ in Eq.~\ref{eq:hadmard} to compute $\mathcal{L}_{sp+}$ for $c\in S $.  Hence, the  gradients ($\delta^{c}$) for the $c^{th}$ channel of the last convolution layer of the density branch, due to $\mathcal{L}_{sp+}$ , are  computed as, 
  \vspace{-0.1cm}
\begin{equation} 
    \delta_{sp+}^c= \pd{\mathcal{\hat{L}}_{sp+}}{\tilde{\textbf{D}}^{c} } \odot \textbf{B}^{c}. 
     \vspace{-0.1cm}
\end{equation}
Since   $\mathcal{L}_{MSE}$, $\mathcal{L}_{rank}$ and $\mathcal{L}_{sp-}$   are computed using  MSE, ranking and logistic BCE losses on convolution outputs, their respective gradients  are computed using off-the-shelf pytorch implementation \cite{pytorch_cite}.

\noindent\textbf{Inference:} 
The image classification branch outputs a class confidence score $s^c$ for each class, indicating the  presence (~$\hat{t_c}>0$, if $s^c>0$) or absence ($\hat{t_c}=0$, if  $s^c \le 0$ ) of  the object category $c$.  
The predicted count $\hat{t_c}$ is obtained by summing the density map $\textbf{D}^c$ for  category $c$ over its entire spatial region. 
The proposed approach only utilizes subitizing annotations ($t_c\le 4$) and accurately predicts object counts for \emph{both} within and beyond subitizing range  (see Fig. \ref{fig:CountVsCountErr}).  
 \subsection{Image-level Supervised Instance Segmentation}
 \label{sec:instanceSeg}
The proposed ILC supervised density map estimation approach can also be utilized for instance segmentation. Note that the local summation of an ideal density map over a ground-truth segmentation mask is one. 
 We use this property to improve state-of-the-art image-level supervised instance segmentation (PRM) \cite{PRM}. PRM employs a scoring metric that combines instance level cues from peak response maps $R$, class aware information from object category maps and spatial continuity priors from off-the-shelf object proposals \cite{mcg_2017,cob_eccv2016}. Here, the peak response maps are generated from local maxima (peaks of $\tilde{\textbf{M}}^{c}$) through a peak back-propagation process \cite{PRM}. The scoring metric is then used to rank object proposals corresponding to each peak for instance mask prediction. We improve the scoring metric by introducing an additional term $d_p$ in the  metric.  The term  $d_p$ penalizes an object proposal $P_r$, if the predicted count in those regions of the density map $\textbf{D}^c$ is different from one, as $d_p$= $|1-\|\textbf{D}^c\cdot P_r\|_{sum}|$. Here, $|~|$ is the absolute value operator. 
 For each peak, the new scoring metric  $Score$ selects the  highest scoring object proposal $P_r$. 
  \vspace{-0.1cm}
\begin{equation} 
\label{eq:score}
    Score=\alpha\cdot R*P_r+R*\hat{P_r}-\beta\cdot Q*P_r-\gamma\cdot d_p.
     \vspace{-0.1cm}
\end{equation}
Here, the background mask $Q$ is derived from object category map and $\hat{P_r}$ is the contour mask of the proposal $P_r$ derived using morphological gradient \cite{PRM}. Parameters $\alpha$,  $\beta$  and $\gamma$ are empirically set as in \cite{PRM}.

 

\vspace{-0.1cm}
\section{Experiments}
\vspace{-0.1cm}
\noindent\textbf{Implementation details:}
Throughout our experiments, we fix the training parameters. An initial learning rate of $10^{-4}$ is used for the pre-trained ResNet-50 backbone, while image classification and density branches are trained with an initial learning rate of  $0.01$. The number of input channels $P$ of $1 \times 1 $ convolution for each branch is set to $P=1.5  \times C $. A mini-batch size of 16 is used for the SGD optimizer. 
The momentum  is set to 0.9 and weight decay to $10^{-4}$. Considering high imbalance between  non-zero and zero counts in COCO dataset (\eg 79 negative categories for each positive category), only 10\% of  samples in the set $A$ are used to train the density branch. 
Code will be made public upon publication. 

\noindent\textbf{Datasets:}
We evaluate common object counting on the PASCAL VOC 2007 \cite{pascal_2012} and COCO \cite{coco_eccv2014} datasets. For fair comparison, we employ same splits, named as count-train, count-val and count-test, as used in the state-of-the-art methods \cite{WhereAreBlobsECCV18},\cite{Chattopadhyay_2017_CVPR}. 
For COCO dataset, the training set is used as count-train, first half of the validation set as the count-val and its second half as the count-test. 
In Pascal VOC 2007 dataset, we evaluated against the count of  non-difficult instances in the count-test as in  \cite{WhereAreBlobsECCV18}. 
For instance segmentation, we train and report the results on the PASCAL VOC 2012 dataset similar to \cite{PRM}.\\
 \noindent\textbf{Evaluation Criteria:}
The predicted count $\hat{t_c}$  is rounded to the nearest integer. We evaluate common object counting, as in \cite{Chattopadhyay_2017_CVPR, WhereAreBlobsECCV18}, using root squared error (RMSE) metric and its three variants namely RMSE non-zero (RMSE-nz), relative RMSE (relRMSE) and relative RMSE non-zero (relRMSE-nz). The $RMSE_c$ and  $relRMSE_c$ errors for category $c$ are computed as $\sqrt{\frac{1}{T}\sum_{i=1}^{T}(t_{ic}-\hat{t_{ic}})^2}$ and,  $\sqrt{{\frac{1}{T}\sum_{i=1}^{T}\frac{(t_{ic}-\hat{t_{ic}})^2}{t_{ic}+1}}}$ respectively. Here, $T$  is the total number of images in the test set and $\hat{t}_{ic}$, $t_{ic}$ are the predicted and ground-truth counts for image $i$.  The errors are then averaged across all categories to obtain the mRMSE and m-relRMSE on a dataset. The above metrics are also evaluated for ground-truth instances with non-zero counts as mRMSE-nz and m-relRMSE-nz. For all error metrics, smaller numbers indicate better performance. We refer to  \cite{Chattopadhyay_2017_CVPR} for more details. For instance segmentation, the performance is evaluated using Average Best Overlap (ABO) \cite{ABO_2015} and ${mAP}^r$, as in \cite{PRM}. The ${mAP}^r$ is computed with intersection over union (IoU) thresholds of 0.25, 0.5 and 0.75.

 

 \noindent\textbf{Supervision Levels:} The level of supervision is indicated as SV in Tab.  \ref{tab:counting_pascal} and \ref{tab:counting_coco}. BB indicates bounding box supervision and PL indicates point-level supervision for each object instance. Image-level supervised methods using only within subitizing range counts are denoted as ILC, while the methods using both within and beyond subitizing range counts are indicated as IC. 

\begin{table}[t]
\resizebox{\columnwidth}{!}{
\begin{tabular}{>{\centering\arraybackslash}p{3cm}|c|cccc}
\hline 
Approach & SV & mRMSE & mRMSE-nz & m-relRMSE & m-relRMSE-nz \\\hline

CAM+regression& IC&	0.45&	1.52&	0.29&	0.64\\\hline
Peak+regression&IC	&0.64	&2.51&	0.30&	1.06\\\hline  \hline
Proposed & ILC & \textbf{0.29} & \textbf{1.14} & \textbf{0.17} & \textbf{0.61}\\\hline
\end{tabular}
}
\vspace{-0.1cm}
\caption{Counting performance on the Pascal VOC 2007 count-test set using our approach and two baselines. Both baselines are obtained by training the network using the MSE loss function.} 
\label{tab:counting_pascal_baseline}
\vspace{-0.10cm}
\end{table}
\vspace{-0.05cm}
\subsection{Common Object Counting Results}
\begin{figure}[t]
		\centering
						\includegraphics[width=0.98\linewidth, clip=true, trim=0cm 15cm 11.5cm 0cm]{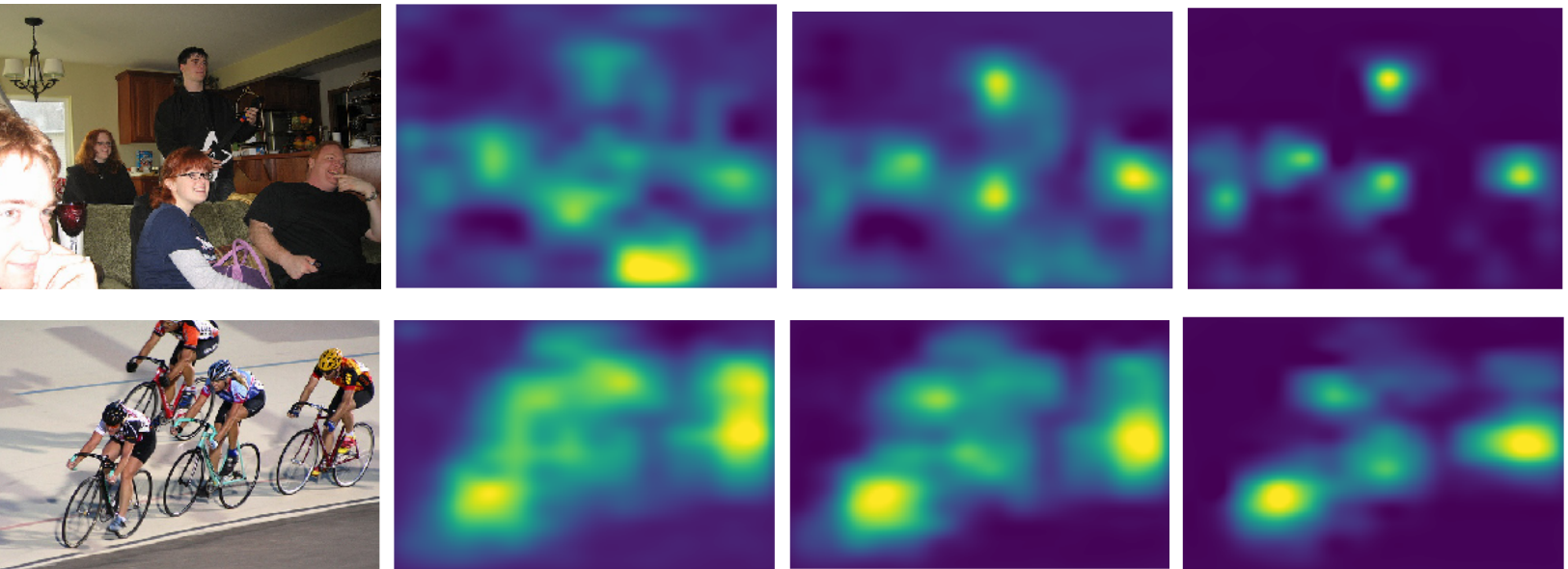}\\ \vspace{-0.1cm}
						\resizebox{\linewidth}{!}{\hspace*{0.06\linewidth} (a) Input Image \hspace*{0.1\linewidth} (b) Class+MSE \hspace*{0.16\linewidth}(c) +Spatial \hspace*{0.15\linewidth} (d) +Ranking  \hspace*{0.08\linewidth} }\\	
				\vspace*{-0.0cm}
			\caption{Progressive improvement in density map quality with the incremental introduction of spatial and ranking loss terms. In both cases (top row: person and bottom row: bicycle), our overall loss function integrating all three terms provides the best density maps. The global object count is accurately predicted (top row: 5 persons and bottom row: 4 bicycles) by accumulation of the respective density map.   
			}
			\label{Fig:experi_densitymap}
			\vspace*{-0.2cm}
\end{figure}
\vspace{-0.1cm}
\label{expe:everyday object counting results}
 \noindent\textbf{Ablation Study:}
We perform an ablation study on the PASCAL VOC 2007 count-test. First, the impact of our two-branch architecture is analyzed by comparing it with two baselines: class-activation \cite{CAM} based regression (CAM+regression) and peak-based regression (Peak+regression) using the local-maximum boosting approach of \cite{PRM}. Both baselines are obtained by end-to-end training of the network, employing the same backbone, using MSE loss function to directly predict global count. Tab. \ref{tab:counting_pascal_baseline} shows the comparison. Our approach largely outperforms both baseline highlighting the importance of having a two-branch architecture with explicit terms in the loss function to preserve the spatial distribution of objects. Next, we evaluate the contribution of each term in our loss function towards the final count performance.

 Fig.~\ref{Fig:experi_densitymap} shows the systematic improvement in density maps (top row: person and bottom row: bicycle) quality with the incremental addition of (c) spatial $\mathcal{L}_{spatial}$  and (d) ranking ($\mathcal{L}_{rank}$) loss terms to the (b) MSE ($\mathcal{L}_{rank}$) loss term. Similar to CAM, the density branch trained with MSE loss alone gives coarse location of object instances. However, many background pixels are identified as part of the object (false positives) resulting in inaccurate spatial distribution of object instances. Further, this inclusion of false positives prevents the delineation of multiple object instances. Incorporating the spatial loss term improves the spatial distribution of objects in both density maps. The density maps are further improved by the incorporation of the ranking term that penalizes the under-estimation of count beyond the subitizing range (top row) in the loss function. Moreover, it also helps to reduce the false positives within the subitizing range (bottom row). Tab. \ref{tab:loss_analysis} shows the systematic improvement, in terms of mRMSE and mRMSE-nz, when integrating different terms in our loss function. The best results are obtained when integrating all three terms (classification, spatial and global) in our loss function. We also evaluate the influence of $\lambda$ that controls the relative weight of the ranking loss. We observe $\lambda=0.1$ provides the best results and fix it for all datasets. 
    \begin{table}[t]
\centering
\resizebox{\columnwidth}{!}{
\begin{tabular}{>{\centering\arraybackslash}c|ccc|p{1mm}|cccc}
\hline
         & \begin{tabular}[c]{@{}c@{}}${\cal{L}}_{class}+$\\ $\mathcal{L}_{MSE}$\end{tabular} & \begin{tabular}[c]{@{}c@{}}${\cal{L}}_{class}+$\\ ${\cal{L}}_{spatial}$\\ +$\mathcal{L}_{MSE}$\end{tabular} &  \begin{tabular}[c]{@{}c@{}}${\cal{L}}$\\ $\lambda=0.1$\\ \end{tabular} && \begin{tabular}[c]{@{}c@{}}${\cal{L}}$\\ $\lambda=0.01$\\ \end{tabular} &  \begin{tabular}[c]{@{}c@{}}${\cal{L}}$\\ $\lambda=0.05$\\ \end{tabular} &  \begin{tabular}[c]{@{}c@{}}${\cal{L}}$\\ $\lambda=0.5$\\ \end{tabular} & \begin{tabular}[c]{@{}c@{}}${\cal{L}}$\\ $\lambda=1$\\ \end{tabular} \\ \cline{1-4} \cline{6-9}
mRMSE    & 0.36  & 0.33 & \textbf{0.29}  & & 0.31& 0.30 & 0.32& 0.36   \\ \cline{1-4} \cline{6-9}
mRMSE-nz & 1.52 & 1.32  & \textbf{1.14} &  & 1.27 & 1.16& 1.23  & 1.40 \\ \hline
\end{tabular}}\vspace{-0.0cm}
\caption{Left: Progressive integration of different terms in loss function and its impact on the final counting performance on the PASCAL VOC count-test set. Right: influence of the weight ($\lambda$)  of ranking loss.}
\label{tab:loss_analysis}
\end{table}

\begin{table}[t]
\resizebox{\columnwidth}{!}{
\begin{tabular}{>{\centering\arraybackslash}p{3cm}|c|cccc}
\hline 
Approach & SV & mRMSE & mRMSE-nz & m-relRMSE & m-relRMSE-nz \\\hline
Aso-sub-ft-3$\times$3 \cite{Chattopadhyay_2017_CVPR} & BB & 0.43 & 1.65 & 0.22 & 0.68 \\\hline
Seq-sub-ft-3$\times$3 \cite{Chattopadhyay_2017_CVPR} & BB & 0.42 & 1.65 & 0.21 & 0.68 \\\hline
ens \cite{Chattopadhyay_2017_CVPR} & BB & 0.42 & 1.68 & 0.20 & 0.65 \\\hline
Fast-RCNN \cite{Chattopadhyay_2017_CVPR} & BB & 0.50 & 1.92 & 0.26 & 0.85 \\\hline
LC-ResFCN \cite{WhereAreBlobsECCV18} & PL & 0.31 & 1.20 & 0.17 & 0.61 \\\hline
LC-PSPNet \cite{WhereAreBlobsECCV18} & PL & 0.35 & 1.32 & 0.20 & 0.70 \\\hline
glance-noft-2L \cite{Chattopadhyay_2017_CVPR} & IC & 0.50 & 1.83 & 0.27 & 0.73 \\\hline
Proposed & ILC & \textbf{0.29} & \textbf{1.14} & \textbf{0.17} & \textbf{0.61}\\\hline
\end{tabular}
}
\caption{State-of-the-art counting performance comparison on the Pascal VOC 2007 count-test. Our ILC supervised approach outperforms existing methods.}
\label{tab:counting_pascal}
\end{table}

\noindent\textbf{State-of-the-art Comparison:}
Tab. \ref{tab:counting_pascal} and \ref{tab:counting_coco} show state-of-the-art comparisons for common object counting on the PASCAL VOC 2007 and COCO datasets respectively.  On the PASCAL VOC 2007 dataset (Tab.~\ref{tab:counting_pascal}), the glancing approach (glance-noft-2L) of \cite{Chattopadhyay_2017_CVPR} using image-level supervision both within and beyond the subitizing range (IC) achieves mRMSE score of $0.50$. Our ILC supervised approach considerably outperforms the glance-noft-2L method with a absolute gain of 21\% in mRMSE. Furthermore, our approach  achieves consistent improvements on all error metrics, compared to state-of-the-art point-level and bounding box based supervised methods.

\begin{table}[t]

\resizebox{\columnwidth}{!}{
\begin{tabular}{>{\centering\arraybackslash}p{3cm}|c|cccc}
\hline 
Approach & SV & mRMSE & mRMSE-nz & m-relRMSE & m-relRMSE-nz \\\hline
Aso-sub-ft-3$\times$3 \cite{Chattopadhyay_2017_CVPR} & BB & 0.38 & 2.08 & 0.24 & 0.87 \\\hline
Seq-sub-ft-3$\times$3 \cite{Chattopadhyay_2017_CVPR} & BB & 0.35 & 1.96 & 0.18 & 0.82 \\\hline
ens \cite{Chattopadhyay_2017_CVPR} & BB & 0.36 & 1.98 & 0.18 & \textbf{0.81} \\\hline
Fast-RCNN \cite{Chattopadhyay_2017_CVPR} & BB & 0.49 & 2.78 & 0.20 & 1.13 \\\hline
LC-ResFCN \cite{WhereAreBlobsECCV18} & PL & 0.38 & 2.20 & 0.19 & 0.99 \\\hline
glance-ft-2L \cite{Chattopadhyay_2017_CVPR} & IC & 0.42 & 2.25 & 0.23 & 0.91 \\\hline
Proposed & ILC & \textbf{0.34} & \textbf{1.89} & \textbf{0.18} & 0.84\\\hline
\end{tabular}
}
\caption{State-of-the-art counting performance comparison on the COCO count-test set. Despite using reduced supervision, our approach provides superior results compared to existing methods on three metrics. 
Compared to the image-level count (IC) supervised approach \cite{Chattopadhyay_2017_CVPR}, our method achieves an absolute gain of 8\% in terms of mRMSE.}
\label{tab:counting_coco}
\end{table}
      \begin{figure}[t]
		\centering
						\includegraphics[width=1\linewidth, clip=true, trim=0cm 14.6cm 3.5cm 0cm]{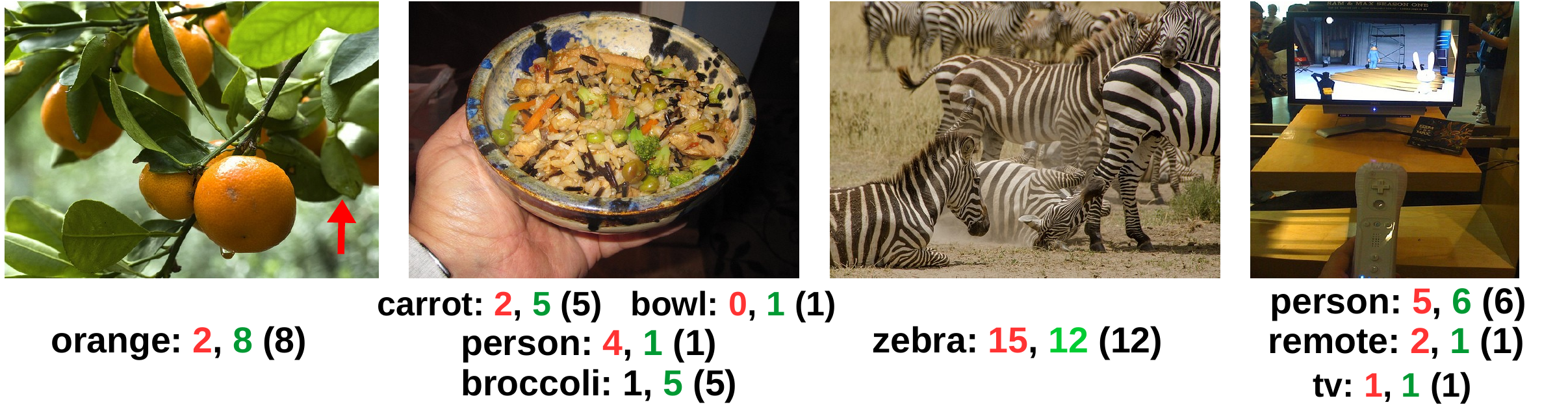}\\ 
			\caption{Object counting examples on the COCO dataset. The ground-truth, point-level supervised counts \cite{WhereAreBlobsECCV18} and our predictions are shown in black, red and green respectively. Our approach accurately performs counting beyond the subitizing range and on diverse categories (fruits to animals) under heavy occlusions (highlighted by a red arrow in the left image).}
			\label{Fig:qual_coco}
		\vspace*{-0.2cm}
\end{figure}

\begin{figure}[t]
	\includegraphics[width=0.95\linewidth, clip=true, trim=0.1cm 11.2cm 14.5cm 0.1cm]{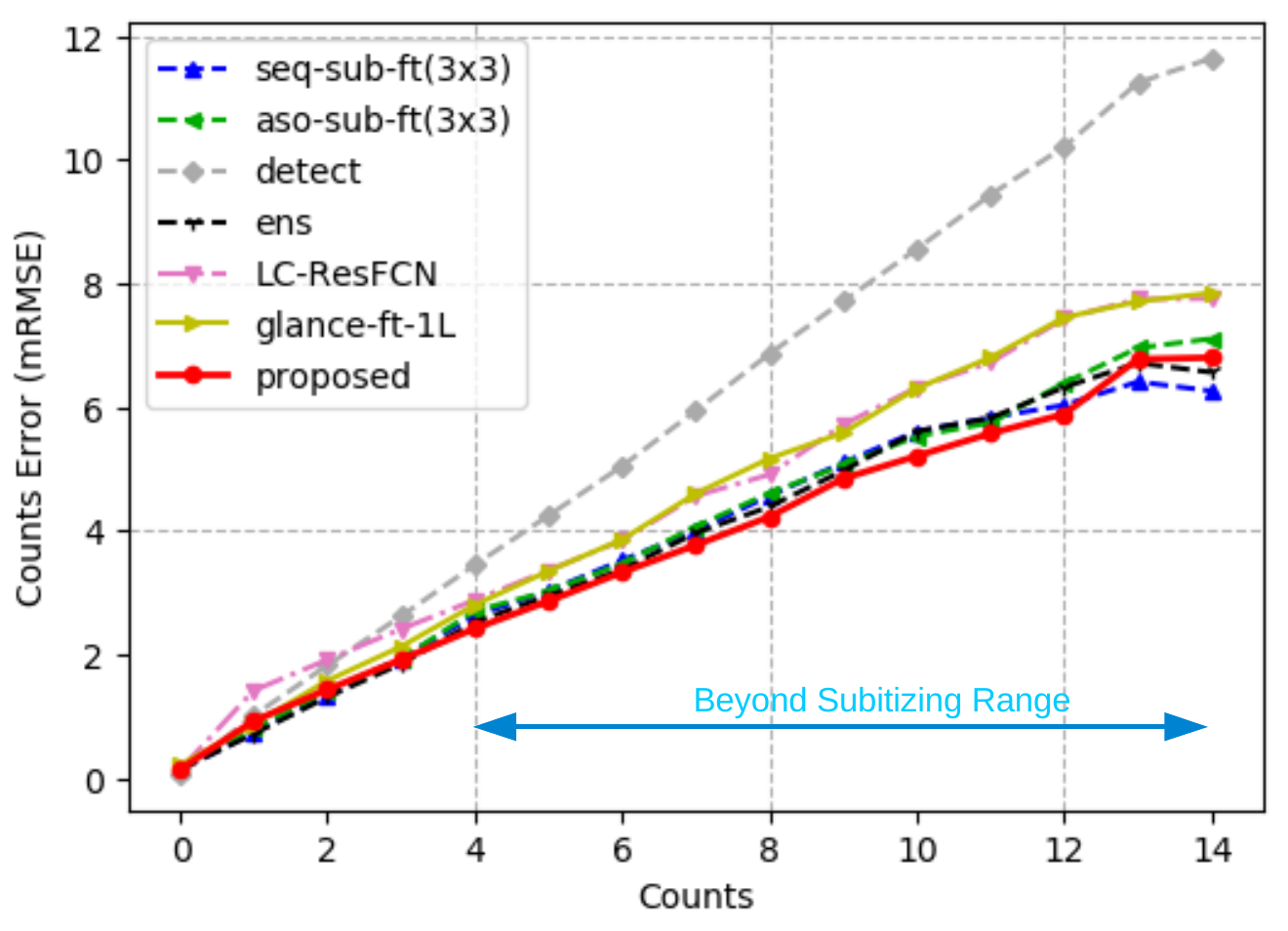}
	\vspace{-0.2cm}
	\caption{Counting performance comparison in RMSE, across all categories, at different ground-truth count values on the COCO count-test set. Different methods, including BB and PL supervision, are shown in the legend. Our ILC supervised approach provides superior results compared to the image-level supervised glancing method. Furthermore, our approach performs favourably compared to other methods using instance-level supervision. }
	\label{fig:CountVsCountErr}
	 \vspace*{-0.3cm}
\end{figure}

Tab. \ref{tab:counting_coco} shows the results on COCO dataset. Among the existing methods, the two BB supervised approaches (Seq-sub-ft-3x3 and ens) yields mRMSE scores of $0.35$ and $0.36$ respectively. The PL supervised LC-ResFCN approach \cite{WhereAreBlobsECCV18} achieves mRMSE score of $0.38$. The IC supervised glancing approach (glance-noft-2L) obtains mRMSE score of $0.42$. Our approach outperforms the glancing approach with an absolute gain of 8\% in mRMSE. Furthermore, our approach  also provides consistent improvements over the glancing approach in the other three error metrics and is only below the two BB supervised methods (Seq-sub-ft3x3 and ens) in m-relRMSE-nz. Fig. \ref{Fig:qual_coco} shows object counting examples using our approach and the point-level (PL) supervised method \cite{WhereAreBlobsECCV18}. Our approach performs accurate counting on various categories (fruits to animals) under heavy occlusions. Fig. \ref{fig:CountVsCountErr} shows counting performance comparison in terms of RMSE, across all categories, on COCO count-test. The x-axis shows different ground-truth count values. We compare with the different IC, BB and PL supervised methods \cite{Chattopadhyay_2017_CVPR, WhereAreBlobsECCV18}. Our approach achieves superior results on all count values compared to glancing method \cite{Chattopadhyay_2017_CVPR} despite not using the beyond subitizing range annotations during training.  Furthermore, we perform favourably compared to other methods using higher supervision.

\noindent\textbf{Evaluation of density map:}
We employ a standard grid average mean absolute error (GAME) evaluation metric \cite{GAME} used in crowd counting to evaluate spatial distribution consistency in the density map. In GAME(n), an image is divided into $4^n$ non-overlapping grid cells. Mean absolute error (MAE) between the predicted  and the  ground-truth local counts are reported for  $n=0,~1,~2$ and $3$, as in \cite{GAME}.  We compare our approach with the state-of-the-art PL supervised counting approach (LCFCN) \cite{WhereAreBlobsECCV18} on the 20 categories of the PASCAL VOC 2007 count-test set. Furthermore, we also compare with recent crowd counting approach (CSRnet) \cite{CSRnetDialatedConv_2018_CVPR} on the person category of the PASCAL VOC 2007 by retraining it on the dataset. For the person category, the PL supervised LCFCN and CSRnet approaches achieve scores of $2.80$ and $2.44$ in GAME(3).
The proposed method outperforms LCFCN and CSRnet in GAME (3) with score of $1.83$, demonstrating the capabilities of our approach in the precise spatial distribution of object counts. Moreover, our method outperforms LCFCN for all 20 categories.

 \subsection{Image-level supervised Instance segmentation}
 \vspace{-0.15cm}
 Finally, we evaluate the effectiveness of our density map to improve the state-of-the-art image-level supervised instance segmentation approach (PRM)  \cite{PRM}  on the PASCAL VOC 2012  dataset (see Sec.~\ref{sec:instanceSeg}). 
  In addition to PRM, the image-level  supervised object detection methods MELM \cite{melm_18}, CAM \cite{CAM} and SPN \cite{spn_iccv2017} used with  MCG mask  and reported by \cite{PRM} are also included in 
 Tab.~\ref{tab:ins_seg_pascal}. 
 
\begin{figure}[t]
		\centering 
			\includegraphics[width=1\linewidth, keepaspectratio,clip=true, trim=0cm 15.2cm 15.5cm 0cm]{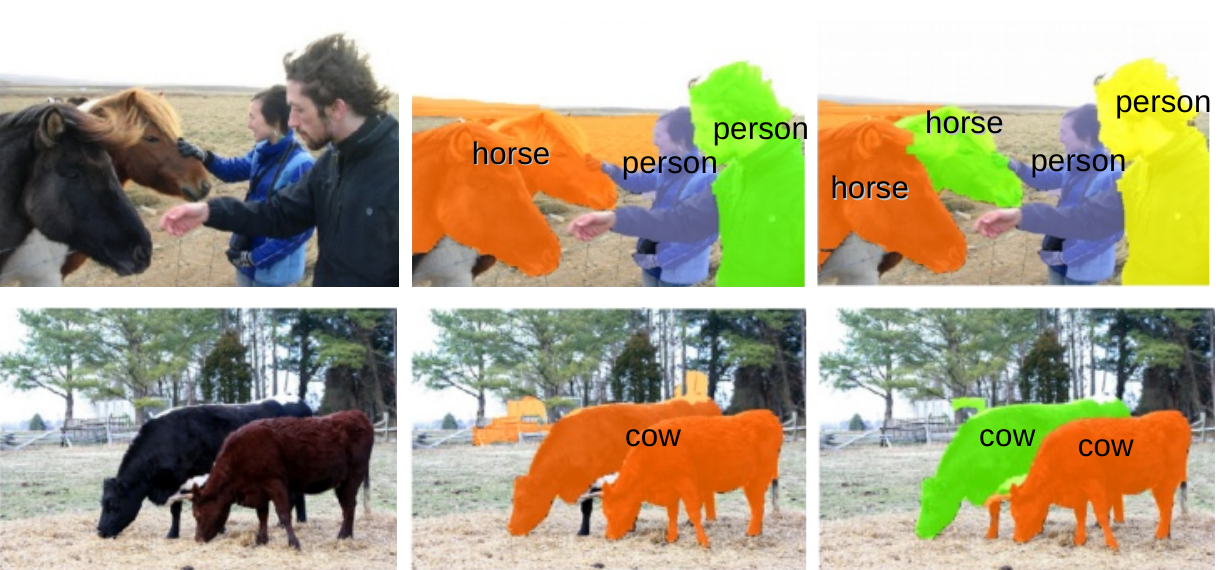}\\
			\hspace*{0.0\linewidth} (a) Input Image  \hspace*{0.07\linewidth} (b) PRM\cite{PRM} \hspace*{0.07\linewidth} (c) Our Approach  \\  
			\caption{Instance segmentation examples obtained using PRM \cite{PRM} and our approach. The proposed approach accurately delineates spatially adjacent multiple object instances of horse and cow categories.}
			\label{fig:experi_instanceSeg}
\end{figure}

 
 The proposed method largely outperforms all the baseline approaches and \cite{PRM}, in all four evaluation metrics. Even though our approach marginally increases the level of supervision (lower-count information), it improves the state-of-the-art PRM with a relative gain of 17.8$\%$ in terms of average best overlap (ABO). Compared to PRM, the gain obtained at lower IoU threshold (0.25) highlights the improved location prediction capabilities of the proposed method. Furthermore, the gain obtained at higher IoU threshold (0.75), indicates the effectiveness of the proposed scoring function in assigning higher scores to the object proposal that has highest overlap with the ground-truth object, as indicated by the improved ABO performance. Fig.~\ref{fig:experi_instanceSeg} shows qualitative instance segmentation comparison between our approach and PRM. 

\begin{table}[t]
\small
 \centering
\resizebox{0.95\columnwidth}{!}{
\begin{tabular}{>{\centering\arraybackslash}p{3cm}|ccccc}
\hline 
\multicolumn{2}{c|}{Method}& \multicolumn{1}{c}{$mAP^r_{0.25}$} & \multicolumn{1}{c}{$mAP^r_{0.5}$} & \multicolumn{1}{c}{$mAP^r_{0.75}$} & \multicolumn{1}{c}{ABO} \\\hline 
\multicolumn{2}{c|}{MELM+MCG \cite{melm_18}} & 36.9 & 22.9 & 8.4 & 32.9 
                              \\\hline
\multicolumn{2}{c|}{CAM+MCG \cite{CAM}} & 20.4 & 7.8 & 2.5 & 23.0 
                              \\\hline
\multicolumn{2}{c|}{SPN+MCG \cite{spn_iccv2017}} & 26.4 & 12.7 & 4.4 & 27.1 
                              \\\hline
\multicolumn{2}{c|}{PRM \cite{PRM}} & 44.3 & 26.8 & 9.0 & 37.6 \\\hline
\multicolumn{2}{c|}{Ours} & \textbf{48.5} & \textbf{30.2} & \textbf{14.4} & \textbf{44.3} \\\hline
\end{tabular}
}
 \caption{Image-level supervised instance segmentation results on the PASCAL VOC 2012 val. set in terms of mean average precision (mAP\%) and Average Best Overlap(ABO). Our approach ourperforms the state-of-the-art PRM \cite{PRM} with a relative gain of 17.8$\%$ in terms of ABO.}
\label{tab:ins_seg_pascal}
\vspace{-0.0cm}
\end{table}
\vspace{-0.2cm}
\section{Conclusion}
\vspace{-0.1cm}
We proposed an ILC  supervised density map estimation approach for common object counting in natural scenes. Different to existing methods, our approach provides both the global object count and the spatial distribution of object instances with the help of a novel loss function.  We further demonstrated the applicability of the proposed density map in instance segmentation. Our approach outperforms existing methods for both common object counting and image-level supervised instance segmentation.

{\small
\bibliographystyle{ieee}
\bibliography{main}
}

\end{document}